# Implementation of Augmented Reality in Autonomous Warehouses: Challenges and Opportunities

David Puljiz
Karlsruhe Institute of Technology, IAR-IPR
Karlsruhe, Germany
david.puljiz@kit.edu

Gleb Gorbachev
Karlsruhe Institute of Technology, IAR-IPR
Karlsruhe, Germany
gleb_gorbachev_2003@mail.ru

Björn Hein
Karlsruhe Institute of Technology, IAR-IPR
Karlsruhe, Germany
bjoern.hein@kit.edu

## ABSTRACT

Autonomous warehouses with mobile, rack-carrying robots are starting to become commonplace, with systems such as Amazon's Kiva and Swisslog's CarryPick already implemented in functional warehouses. Such warehouses however still require human intervention for object picking and maintenance. In the European project SafeLog we are developing a safety-vest, used for safety-critical ranging and stopping of mobile robots, an improved planner that can handle large fleets of heterogeneous agents as well as an AR interaction system to navigate and support human workers in such automated environments. Here we present the AR interaction modalities, namely navigation, pick-by-AR and general system interactions that were developed at the moment of writing, as well as the overall system concept and planned future work.

## 1 INTRODUCTION

Automating intralogistics with mobile robots has been a major research direction for more than a decade now, with the KIVA system being the first widely adopted system of this kind [21]. Many other such systems have been developed and are being developed since then including Swisslog's CarryPick [17], Scallog's Scallog system [16], Fetch Robotics' Freight [15] etc. Although autonomous picking from racks has seen major progress in recent years, in part also due to the Amazon Picking Challenge [4], it has still to become more robust to be implemented into real warehouses. Thus, even though the robots carry the racks, the warehouses themselves are far from being fully automated and still require human presence. The automated rack carrying system however has allowed a big increase in efficiency [6].

In such a system the area containing the racks is off-limits to human workers, usually separated by heavy-duty fences and laser curtains, as prescribed by e.g. [2] [1] Mobile robots navigate in that space, pick-up the racks, and deliver them to the picking stations where human workers pick ordered objects. These robots don't posses any sensor except a camera for detecting markers, which are used for navigation, and perhaps bumpers. The navigation and control are and should be kept simple. This is to keep the cost of the robots low and allow scalability. If a human needs to enter the warehouse the entire fleet of mobile robots needs to stop resulting in significant downtime. There are three main reason why a human worker would need to enter the mobile robot area: repairing a broken robot, pre-picking or going to centrally located picking stations. Broken robots can be quite troublesome, as mobile robots in automated warehouses usually use floor markers to navigate. This means that if the signal to a robot is lost, an entire aisle may be blocked since the exact position of the robot is not precisely known, lowering efficiency and complicating planing. Pre-picking is done when it is inefficient to order a mobile robot to bring an entire rack for picking, instead sending a human worker into a warehouse to pick desired objects. This is impossible to do with current systems and safety guidelines. Finally, by placing the picking stations in the center of the warehouse, travel time to and from the stations could be greatly reduced. This would however also require planing for forklifts (autonomous or manned) and other vehicles that would ferry packages from the station to the delivery vehicles outside the warehouse.

In the future warehouse, where humans and mobile robots can work side-by-side, augmented reality would play a major role in helping the human workers navigate the ever-changing warehouse environment, assist with picking and generally enable new interaction modalities with the system. Augmented Reality has already been tested and proven in unautomated warehouses in companies such as Intel [20], resulting in increased efficiency. We strongly

---

[1]This is according to the quite strict EU safety guidelines, the KIVA system for example just separates the workspaces with a line on the floor

believe that coupling the native efficiency gains of automated warehouses and AR could produce a warehouse more efficient than the sum of it's parts.

The rest of the paper is organized as follows: The next section presents the general system architecture of automated warehouses as well as the new system components developed by the project with the emphasis on the interaction modalities offered by AR in such a setting. Section 3 describes the elements already implemented. In Section 4 we go over the future work and the possible challenges. Section 5 gives a recap and some concluding remarks. The paper concludes with acknowledgements.

## 2 SYSTEM DESCRIPTION

The main parts of a standard automated warehouse are the Warehouse Management System (WMS) and the Fleet Management System (FMS). The WMS takes care about the orders - which product needs to be be delivered as well as when and where. It has a database which contains information about the location of products in racks, thus it knows precisely which rack contains the wanted product. The WMS gives commands to the FMS in the form of which rack should be brought to which picking station and at what time. The FMS then coordinates the fleet of mobile robots to try and fulfill all of the commands.

The project SafeLog implements an improved FMS system that can work with bigger, heterogeneous fleets. In most cases this will include mobile robots and human workers, however manned or unmanned forklifts or other vehicles can also be managed. The planner also takes care that no mobile robots come into vicinity of human workers. A planner however is incapable of being safety certifiable, mostly due to the fact that the computing power and libraries required by such a system would require safety certificates on every single component. This is why a safety vest was also developed to be safety certifiable and take care of the ranging and provide a final safety barrier in case the planer fails. If a mobile robot comes within the first safety radius (8-10 m) it is slowed down. If it comes within 8 m it is stopped completely. The safety radius can of course be increased or decreased. The mobile robots themselves navigate using ground markers usually 0.5m apart. These ground nodes can be thought of as vertices on a graph and are useful both for localization and for rendering the path in AR. These systems are crucial to understanding the limits and possibilities of AR interactions in automated warehouses. An example of an automated warehouse layout can be seen in Figure 1.

Regarding the components used, the automated warehouse components are part of the CarryPick system provided by our partner Swisslog. For AR interaction we use the Microsoft Hololens [9] due to the (so far) robust localization and the fact that it's untethered. For the programing environment Unity 2017.1.1f [19] is used together with Microsoft's Mixed Reality Toolkit [10] and Vuforia [5].

As mentioned before, the AR interaction with the system can be separated into four main categories: assistance during navigation of the warehouse, assistance for object picking, assistance for robot repair and other system interactions. For object picking and robot repair we decided that having both hands free using an AR headset (instead of e.g. having a tablet) is especially important. The basic

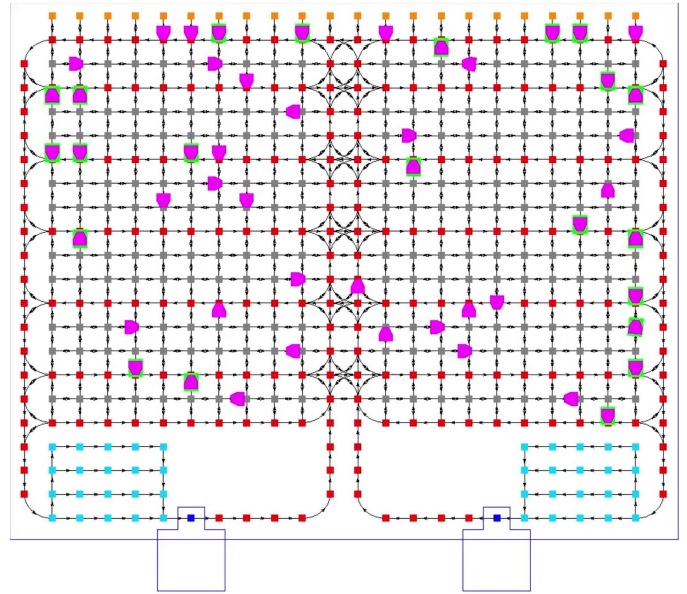

**Figure 1: An example of an autonomous warehouses taken from the simulator. Red squares are navigation nodes, gray squares are rack position nodes, yellow squares are charge/maintenance nodes, light blue squares are queue nodes, dark blue squares are picking nodes. The dark blue lines around the picking nodes symbolize the picking stations. Robots are shown in pink while robots carrying racks have a green polygon around them. The large blue square indicates the no-entry zone for human workers. This illustrates the sort of environment the worker needs to navigate in**

navigation modalities could be implemented using a standard tablet even without AR with a map that was constantly updated. This however would provide less interaction modalities and require the worker to look at the tablet constantly, reducing situational awareness and comfort. Additionally the platform would need to be changed for different interaction modalities. It was therefore decided to implement everything on a single AR headset platform.

### 2.1 Navigation Assistance and localization

Large unautomated warehouses can be difficult to navigate and automated ones, which aren't designed to be easily navigated by people, even more so. Perhaps the biggest challenge is that the goal is unknown. In case a human worker is sent to repair a broken robot, the robot needs to be localized. Likewise if a worker enters the warehouse for pre-picking, the position of the racks are not fixed. The same rack can be picked up from one place and dropped off in another, making fixed signs and maps unusable. When a worker needs to enter the warehouse, the FMS updates the position of her or his goal and calculates the path to the goal in such a way that the encounters with mobile robots are non-existent or at least minimized. The path is then shown on the AR glasses using different modalities:

- A line on the floor is shown that guides the worker through the aisles to the goal position. Lines connect the ground nodes and are subject to occlusion and distance culling. This reduces the rendering requirements and gives the worker a more immersive feeling in the case of occlusion culling. The distances to the goal may also be rendered.
- Overhead a particle stream also guides the worker. This is done so that the worker may look up and around without losing the path information
- when the worker is in close proximity of the goal, an arrow points at the direction of the goal.
- Both 2D and 3D minimaps can be opened by the worker showing the live status of the warehouse including current position of racks, mobile robots and other workers in the warehouse.

To make the worker feel more comfortable in an environment where mobile robots carrying 600+ kg payloads may pass within 10-15m from him, we give the worker X-ray vision. By this we mean that robots within a certain radius (e.g. from 10 to 20 meters away) will have their 3D models rendered even through obstacles, as well as their path. This should provide situational awareness to the worker and make the trek through the warehouse more comfortable. The worker may also choose to display the safety radius around her-/himself. Again this is to make the worker aware how close a robot is and improve her or his comfort with the system.

A big challenge here is the localization of the AR device. Although the present-day SLAM algorithms [22] running on most AR devices provide precise and robust localization, the algorithm needs to work in a monotonous, changing environment presenting significant problems to standard algorithms which require stable, distinct features. The problems can be mitigated by detecting the already mentioned ground nodes used for robot navigation as well as updating the environment map on the fly. Since the changes in the environment are predictable and well known this can be done reliably. More about the implementation will be presented in the next two sections. An entry procedure will also be developed to ensure that we know the exact position and orientation of the AR Device at the start. This would also then enable the fallback strategy of using pure Visual-Inertial Odometry [12] combined with ground marker detection [18] instead of the in-built SLAM algorithm.

## 2.2 Pick-by-AR

Once the worker reaches the goal and the goal happens to be picking an object from a rack, the AR-based picking assistance will be activated. The WMS knows the structure of the rack and the (at least rough) position of the object in the rack. The rough position of the object is highlighted on the rack and either an arrow or a line points from the center of the field of view to the center of the highlight. An object description and picture may be invoked for easier recognition which should be especially helpful for cluttered racks. Once the object is picked, the camera on the AR Device performs either object detection if available or barcode reading, to check if it's the right object and to log that the object has been picked. This is being done by our partners in Fraunhofer IML [7]. Although we know the exact position of the rack, the localization of the AR Device may be a bit off. For this reason a rack detection algorithm based on rectilinear structure detection via Gaussian Spheres [1] is being developed by our partners at CVUT. This should localize the rack relative to the AR Device without the need to have precise global localization of the AR Device itself.

The AR Device can also be used at the picking stations. In this can the AR Device would also show where to place the picked object. However the pick-to-light system, consisting usually of an overhead projector lighting up the position of the object to pick, a barcode scanner and a screen giving additional information such as object description, is already being widely used and has proven to be extremely effective. We find it doubtful that an AR-based approach could significantly improve this system, although research by Guo et al. seems to indicate that HUD based methods are indeed faster [3]. However tests will be made and the approaches compared in the future.

## 2.3 Robot repair

Being developed by our partners at Fraunhofer IML [7], the robot repair assistant can guide a worker through diagnostics and repair of a broken mobile robot on the spot. That means that the technicians have access to the technical manual everywhere they go. This would also enable untrained workers in the warehouse to quickly respond and diagnose malfunctioning robots. The camera stream can also be used and an expert queried remotely in case of unforeseen malfunctions.

## 2.4 Other Interaction Modalities

The live camera feed can also be used if the worker inside the warehouse notices some other malfunctions, e.g. objects fallen from the rack that might get caught in the wheels of the mobile robots, misaligned racks etc. The supervisor can then cross reference the object with the appropriate rack by querying the WMS, or send additional workers or robots to help realign the rack. As an additional benefit to having onboard localization, we can perform intention recognition of the workers (human intention recognition - HIR). This is being developed by our partners at the UNIZG-FER [14]. By using Markov chains, the goal of the worker may be predicted. Although most of the time the goal will be the one the worker was given in the first place, this algorithm can for example detect if the user is going to the right goal but got lost, if the user started to go back without picking an object from the rack, or if the user is running towards the exit which would indicate a state of emergency and the mobile robots could be stopped.

In this section we first presented the standard system architecture of autonomous warehouses and the new systems that the SafeLog project is introducing. We then proceeded to explain all of the possible interaction modalities that AR enables, the reasoning behind them, as well as some challenges. In the next section we give a brief overview of the parts of the system already implemented at the moment of writing.

## 3 IMPLEMENTATION

In this section we give an overview of the implemented components which were discussed in the previous section as well as the considerations and challenges presented by implementing the components from the conceptual system design to real hardware and

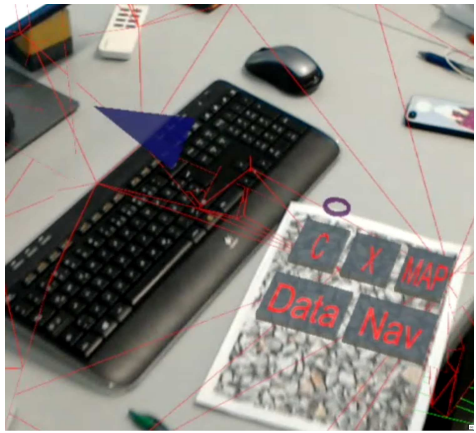

(a) A menu overlayed above a textured image

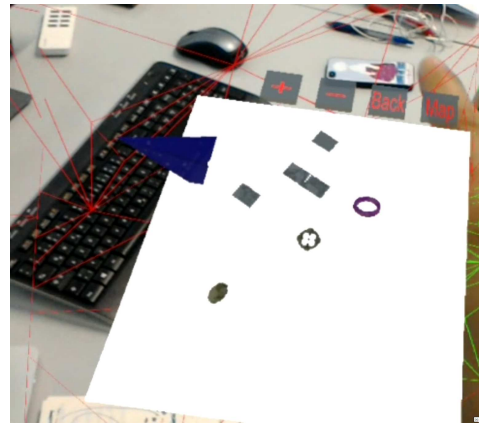

(b) A 2D minimap overlayed above a textured image

**Figure 2: The menu overlayed atop of a highly-texture paper in a) and a 2D minimap showing the user's current position, racks and goal in b). The overlays were done in Vuforia. One can see the overlay has full 6 degrees of freedom and depends on the position and orientation of the paper. The mesh is made visible for debugging purposes**

real world surroundings. The work will be presented in the same order as presented in the previous section.

Implementing line navigation was first attempted using spatial anchors [11], linking the user to the goal also through obstacles since the Hololens isn't connected to any sort of path planner at the moment. Holograms should stay within a 3m radius of the spatial anchor. We at first considered having spatial anchors at every ground node which are 0.5m apart. That would mean that the line segments are 0.5m long and well within spatial anchor limits. However as mentioned in [11], one should avoid making grids of spatial anchors, instead choosing a stationary frame of reference. This is also recommended for dynamic holograms, such as the holographic robots used in the X-Ray vision. The stationary frame of reference can be robustly implemented in our case since we have a structured environment with known entry points. Known entry points are important since the Hololens lacks a global origin point. The origin point is instead taken as the point where the program starts. The worker will have an entry procedure where she or he will need to look at different markers at one of the entry points, which will be scanned using the already mentioned ground node detection algorithm [18]. That way the "true" origin of the world coordinate system can be set which should enable a robust stationary frame of reference. While implementing the line guidance we noticed that the user would be required to look at the floor constantly, so it was decided to implement the particle guidance overhead to prevent such discomfort. The distance to the goal (or in the future the distance to the next turn) was also implemented on the line for better situational awareness. The minimap was first thought to be implemented as head locked content, however head locked content is discouraged for comfort reasons [8]. The problem also arose that the minimap blocks a large portion of the field of view (FoV). The minimap and the other menus are therefore shown as floating menus. This menu can be invoked using a very textured picture on a piece of paper. Vuforia provides options to then overlay holograms over such an image. The hologram has full 6 degrees of freedom.

This can be seen in Figure 2a for an overlayed menu as well as in Figure 2b for an overlayed 2D minimap. The implementation of the line following can be seen in Figure 3.

The X-Ray vision for robots can be seen in Figure 5. This was implemented using full 3D robot models to simulate real robots,

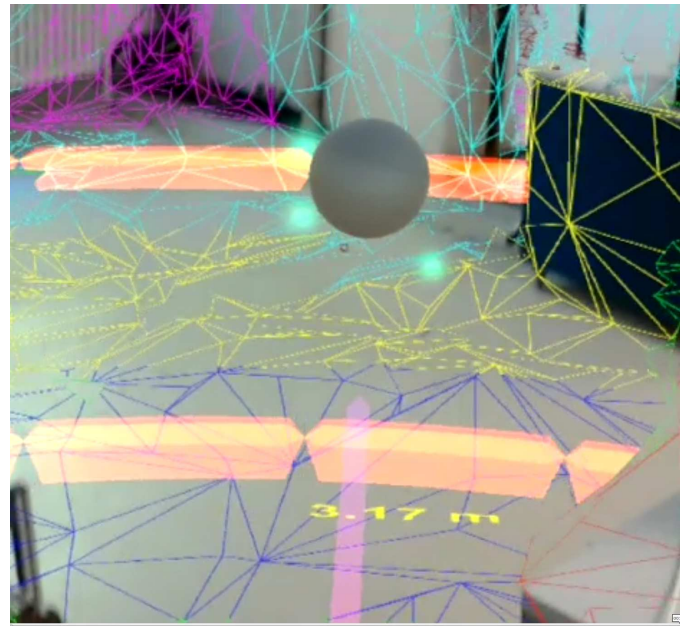

**Figure 3: Line following interaction with the distance to the goal visible. Here one can also see that the safety level display was enabled, with the inner circle corresponding to safety level A where robots need to stop, while the outer one is safety level B where the robots slow down. The mesh is once again made visible for debugging purposes**

while the X-Ray vision was implemented by highlighting the edges of the model. This has proven to work well with occlusions both from real obstacles (Figure 5a) and other holograms (Figure 5b). Implementing spatial sound is also being considered so the worker can also track robots which aren't directly in the FoV.

The first implementation of pick-by-AR can be seen in Figure 4, implemented with a virtual rack. Here head locked content such as a line or an arrow are allowed since the entire interaction only takes place for a few second and the arrow disappears when the center of the FoV is near the highlighted rack area. The highlight basically emulates a pick-to-light system while the arrow tries to minimize unnecessary head movement while looking for the highlighted area. A "game" was implemented where the user is timed for picking a random order of boxes. When implemented or real racks, this will provide good estimation of the efficiency of different interaction modalities for picking.

The object detection for picking and placing help has been implemented by Fraunhofer IML and works well with test objects, however it needs first to be integrated with the pick-by-AR and evaluated for a final conclusion.

The Robot repair developed was also developed by Fraunhofer IML for an in-house robot design to demonstrate the assistance while exchanging a battery for the robot.

The human intention recognition was implemented and tested on a 2D occupancy grid with good results. Details of the implementation and results can be found in [14] by Petković et al.

Among the other implementation considerations perhaps the important is that 3D models should be used sparsely and with the lowest resolution possible as to avoid framedrops. Framedrops are particularly bad in AR and VR applications as they can induce physical discomfort and even dizziness and nausea [13]. The camera stream requires high bandwidth and was usually found to be lagging about 10s no matter the amount of bandwidth available using the Hololens' Device portal. Attention should be given to provide enough bandwidth and care should be given to prevent lag during camera data streaming.

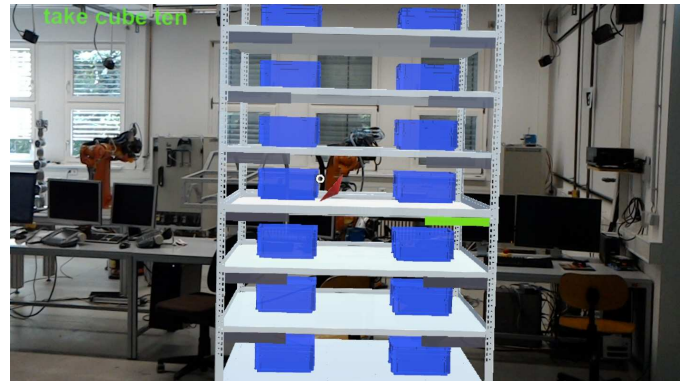

**Figure 4: The first version of pick-by-AR with a virtual rack. One can see the standard Hololens cursor with an added arrow pointing at the object to pick. The area where the object is located is highlighted in green. In the upper left corner is the description of the object. This can also be replaced by an image of the object itself. In this scenario the user is timed while picking objects in a random order to evaluate the speed of different interaction strategies**

## 4 FUTURE WORK

As mentioned in the previous section the vast majority of the interaction modalities have been implemented and evaluated within the project. In case of the navigation and pick-by-AR these interaction modalities need to be implemented with real world object in testable scenarios. The interaction modalities for navigation, pick-by-AR and robot repair can then be evaluated by independent users to evaluate which interaction modality, or the combination thereof gives the user the best experience and efficiency. Once evaluated all of these components should be integrated in a single framework and tested in relevant environments to evaluate the whole system and detect possible conflicts (e.g. the navigation assistance displays the arrow to the goal while at the same time the pick-by-AR displays the arrow towards the object on the rack). The robustness of the rack detection algorithm needs to be tested as well as a robot detection algorithm implemented, since, as mentioned before, when a robot loses connection we cannot be sure about it's exact position, unlike the racks.

Testing should also be done in regards to the possible localization problems and fallback strategies implemented if the localization proves to be lacking in such an environment. This brings us to problem of finding relevant test environments. Users of automated warehouses would experience significant downtime during the tests, with an opportunities to test the system in the actual environment arising only during scheduled maintenance periods. Small test environments do exist and by the time you are reading this the first tests may have already been made. Although evaluated in such a test environment the system may not perform well in full scale warehouses. This is why in parallel we are working on a Virtual Reality test environment for rapid prototyping where AR and other components may be tested and possible problems not replicable in small scale setups detected.

Such an approach will be tested in the coming moths comparing the results of Human Intention Recognition using data from the Hololens and the data from the VR test environment to evaluate how the results change and improve future versions of the algorithm.

In the final section we will give a short recap and the conclusions reached so far during implementation and system design for AR interactions in automated warehouses.

## 5 CONCLUSION

Allowing human workers to freely and safely move through an automated warehouse is the next logical step for warehouse automation. Such a scenario provides numerous challenges especially in regards of supporting the workers in a warehouse environment not designed for humans and difficult to adapt to be both comfortable for humans and simple for the robots to work in. Augmented reality provides a solid alternative where the environment can be made simple for the almost sensorless robots to effectively work, however the workers are still supported and comfortable due to additional information provided by the AR device.

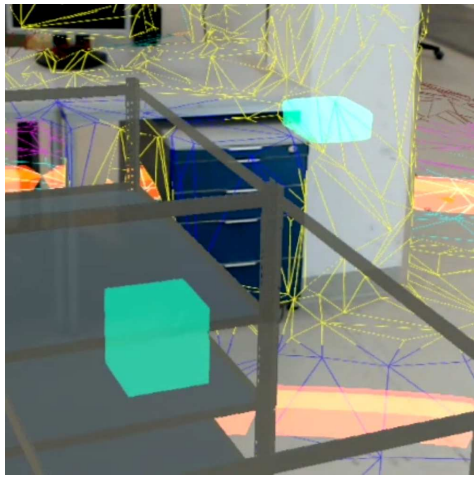

(a) The X-ray vision behind a physical object

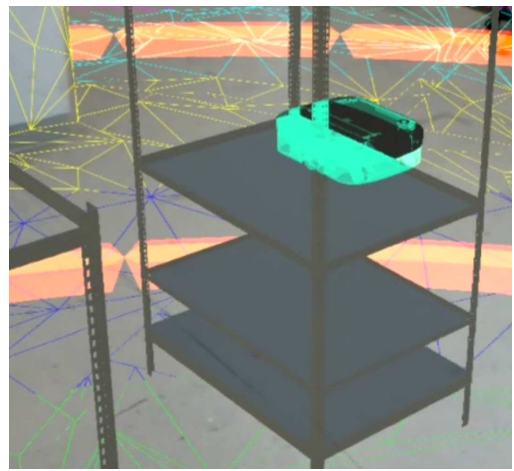

(b) X-Ray vision behind a virtual obstacles

Figure 5: The X-Ray vision can be seen in a) behind real obstacles and in b) behind virtual obstacles. It was achieved by highlighting the edges of the 3D model. The mesh is made visible for debugging purposes

As described in Section 2, navigation, assistance for picking and robot repair as well as other system interactions such as Human Intention Recognition (HRI) all need to be addressed. Navigation in particular is important for an environment that constantly changes. Care should be given to make the human comfortable in an environment with heavy mobile robots passing in the vicinity. This can most easily be done through increased situational awareness. The assistance during picking and robot repair increases the speed and efficiency of the worker in the warehouse. With the live camera feed and communication options enabled by the AR device, quick interventions or even preventions can be made by the workers inside the warehouse with the support of remote specialists. All of this combined should see a large increase in efficiency for such systems.

This approach is not without challenges however, the biggest of which may be the need for robust localization in a changing monotonous environment, where current SLAM techniques may struggle. Secondly, although the field of view of AR devices has seen a steady increase, it still remains relatively small and attention should be given not to overload it with too much information. One should especially be wary of head-locked content since it takes up a lot of the FoV and usually makes the users uncomfortable. Whenever possible information should be outsourced to body-locked floating menus.

We are still at the beginning of the development of AR for the assistance of workers in automated warehouses, however we see great promise in this direction and no doubt, more challenges ahead.

## 6 ACKNOWLEDGMENTS

The work presented in this paper was conducted within the EU project SafeLog - "Safe human-robot interaction in logistic applications for highly flexible warehouses" co-founded by European Union's Horizon 2020 research and innovation program under grant agreement No 688117.

## REFERENCES
[1] Jean-Charles Bazin and Marc Pollefeys. 2012. 3-line ransac for orthogonal vanishing point detection. In *Intelligent Robots and Systems (IROS), 2012 IEEE/RSJ International Conference on*. IEEE, 4282–4287.
[2] Health Great Britain, Safety Executive Staff, Great Britain. Health, and Safety Executive. 2007. *Warehousing and Storage: A Guide to Health and Safety*. HSE Books. https://books.google.de/books?id=HNXMMgEACAAJ
[3] Anhong Guo, Shashank Raghu, Xuwen Xie, Saad Ismail, Xiaohui Luo, Joseph Simoneau, Scott Gilliland, Hannes Baumann, Caleb Southern, and Thad Starner. 2014. A Comparison of Order Picking Assisted by Head-up Display (HUD), Cart-mounted Display (CMD), Light, and Paper Pick List. In *Proceedings of the 2014 ACM International Symposium on Wearable Computers (ISWC '14)*. ACM, New York, NY, USA, 71–78. https://doi.org/10.1145/2634317.2634321
[4] Carlos Hernandez, Mukunda Bharatheesha, Wilson Ko, Hans Gaiser, Jethro Tan, Kanter van Deurzen, Maarten de Vries, Bas Van Mil, Jeff van Egmond, Ruben Burger, Mihai Morariu, Jihong Ju, Xander Gerrmann, Ronald Ensing, Jan Van Frankenhuyzen, and Martijn Wisse. 2017. Team Delft's Robot Winner of the Amazon Picking Challenge 2016. In *RoboCup 2016: Robot World Cup XX*, Sven Behnke, Raymond Sheh, Sanem Sarıel, and Daniel D. Lee (Eds.). Springer International Publishing, Cham, 613–624.
[5] Vuforia / PTC Inc. 2018. Vuforia. https://www.vuforia.com/. (2018). Accessed: 2018-02-09.
[6] Eugene Kim. 2016. Amazon's $775 million deal for robotics company Kiva is starting to look really smart. http:http://www.businessinsider.de/kiva-robots-save-money-for-amazon-2016-6?r=US&IR=T. (2016). Accessed: 2018-02-07.
[7] Thomas Kirks, Jana Jost, and Benedikt Mättig. 2017. AR for Optimization of Processes in Intralogistics. (06 2017). https://www.researchgate.net/publication/320489972_AR_for_Optimization_of_Processes_in_Intralogistics
[8] Microsoft. 2018. Coordinate Systems - Avoid Heal Locked Content. https://developer.microsoft.com/en-us/windows/mixed-reality/coordinate_systems. (2018). Accessed: 2018-02-09.
[9] Microsoft. 2018. Microsoft Hololens. https://www.microsoft.com/en-us/hololens. (2018). Accessed: 2018-02-09.
[10] Microsoft. 2018. Mixed Reality Toolkit. https://github.com/Microsoft/MixedRealityToolkit-Unity. (2018). Accessed: 2018-02-09.
[11] Microsoft. 2018. Spatial Anchors. https://developer.microsoft.com/en-us/windows/mixed-reality/spatial_anchors. (2018). Accessed: 2018-02-09.
[12] D. Nister, O. Naroditsky, and J. Bergen. 2004. Visual odometry. In *Proceedings of the 2004 IEEE Computer Society Conference on Computer Vision and Pattern Recognition, 2004. CVPR 2004.*, Vol. 1. I–652–I–659 Vol.1. https://doi.org/10.1109/CVPR.2004.1315094
[13] Charles M Oman. 1989. Sensory conflict in motion sickness: an observer theory approach. (1989).
[14] Tomislav Petković, Ivan Marković, and Ivan Petrović. 2018. Human Intention Recognition in Flexible Robotized Warehouses Based on Markov Decision Processes. In *ROBOT 2017: Third Iberian Robotics Conference*, Anibal Ollero, Alberto

Sanfeliu, Luis Montano, Nuno Lau, and Carlos Cardeira (Eds.). Springer International Publishing, Cham, 629–640.
[15] Fetch Robotics. 2018. Virtual Conveyor. http://fetchrobotics.com/automated-material-transport-v3/. (2018). Accessed: 2018-02-09.
[16] Scallog. 2018. Scallog System. http://www.scallog.com/en/goods-to-man/. (2018). Accessed: 2018-02-09.
[17] Swisslog. 2018. CarryPick: Flexible and modular storage and order picking system. https://www.swisslog.com/en-us/warehouse-logistics-distribution-center-automation/products-systems-solutions/asrs-automated-storage-,-a-,-retrieval-systems/boxes-cartons-small-parts-items/carrypick-storage-and-picking-system. (2018). Accessed: 2018-02-09.
[18] Jérémy Taquet, Gaël Ecorchard, and Libor Přeučil. 2017. Real-Time Visual Localisation in a Tagged Environment. *arXiv preprint arXiv:1708.02283* (2017).
[19] Unity Technologies. 2018. Unity3D. https://unity3d.com/. (2018). Accessed: 2018-02-09.
[20] Ubimax. 2017. Intel Case Study. http://www.ubimax.com/en/references/intel-casestudy.html. (2017). Accessed: 2018-02-09.
[21] Peter R. Wurman, Raffaello D'Andrea, and Mick Mountz. 2007. Coordinating Hundreds of Cooperative, Autonomous Vehicles in Warehouses. In *Proceedings of the 19th National Conference on Innovative Applications of Artificial Intelligence - Volume 2 (IAAI'07)*. AAAI Press, 1752–1759. http://dl.acm.org/citation.cfm?id=1620113.1620125
[22] Khalid Yousif, Alireza Bab-Hadiashar, and Reza Hoseinnezhad. 2015. An Overview to Visual Odometry and Visual SLAM: Applications to Mobile Robotics. *Intelligent Industrial Systems* 1, 4 (01 Dec 2015), 289–311. https://doi.org/10.1007/s40903-015-0032-7